\documentclass[11pt]{article}

\usepackage[preprint]{acl}

\usepackage{times}
\usepackage{latexsym}

\usepackage[T1]{fontenc}

\usepackage[utf8]{inputenc}

\usepackage{microtype}

\usepackage{inconsolata}

\usepackage{graphicx}

\usepackage{amssymb} 
\usepackage{amsmath} 

\usepackage{graphicx} 
\usepackage{subcaption}

\usepackage{enumitem} 

\usepackage[most]{tcolorbox} 

\newtheorem{proposition}{Proposition}

\usepackage{booktabs}

\usepackage{stfloats} 

\usepackage{colortbl} 

\usepackage{xcolor,pifont}

\usepackage{fontawesome5}

\definecolor{blue1}{rgb}{0.118, 0.251, 0.486}
\definecolor{blue2}{rgb}{0.918, 0.949, 0.996}
\definecolor{lightpurple}{rgb}{0.941, 0.92, 1.000} 
\definecolor{darkpurple3}{rgb}{0.35, 0.35, 0.80}
\definecolor{darkpurple}{rgb}{0.25, 0.25, 0.60}   
\definecolor{normalpurple}{rgb}{0.62, 0.436, 0.706}   
\definecolor{lightorange}{rgb}{1, 0.702, 0.278}
\definecolor{lightorange}{rgb}{1, 0.85, 0.60}

\newtcolorbox{promptbox}[1]{
  colback=blue!5,     
  breakable,                 
  colframe=black!75!black,    
  colbacktitle=black,         
  coltitle=white,             
  title=#1,                   
  boxrule=1pt,                
  arc=1mm,                    
  enhanced,
  attach boxed title to top left={xshift=3mm,yshift=-3mm},
  boxed title style={height=6mm},  
  left=3mm,
  right=3mm,
  top=3.0mm,
  bottom=1.0mm,
  width=0.49\textwidth,
  center,
}

\newtcolorbox{idea}[1]{
  breakable,
  colframe=black!75!black,
  colbacktitle=black,
  coltitle=white,
  title={\faLightbulb\quad #1},
  boxrule=1pt,
  arc=1mm,
  enhanced,
  attach boxed title to top left={xshift=3mm,yshift=-3mm},
  boxed title style={height=6mm},
  left=3mm,
  right=3mm,
  top=3.0mm,
  bottom=1.0mm,
  width=0.49\textwidth,
  center,
}

\title{Are Full Rollouts Necessary for On-Policy Distillation?}

\author{%
  \textbf{Yaocheng Zhang}\thanks{Work done during an internship at Meituan.} \ $^{1,2}$, \ \
  \textbf{Jiajun Chai}$^{3}$, \ \
  \textbf{Yuqian Fu}$^{1,4}$, \ \
  \textbf{Songjun} Tu$^{1,4}$,\ \
  \textbf{Xiaohan Wang}$^{3}$, \\
  \textbf{Wei Lin}$^{3}$, \ \
  \textbf{Guojun Yin}$^{3}$, \ \
  \textbf{Qichao Zhang}$^{1,4}$, \ \
  \textbf{Yuanheng Zhu}\thanks{Corresponding author} \ $^{1,4}$,\ \
  \textbf{Dongbin Zhao}$^{1,2,4}$ \\
  $^{1}$Institute of Automation, Chinese Academy of Sciences \\
  $^{2}$School of Advanced Interdisciplinary Sciences, University of Chinese Academy of Sciences \\
  $^{3}$Meituan \quad
  $^{4}$School of Artificial Intelligence, University of Chinese Academy of Sciences \\
  \texttt{\{zhangyaocheng2023,yuanheng.zhu\}@ia.ac.cn} \\
}

\begin{document}
\maketitle
\begin{abstract}
On-policy distillation (OPD) provides dense teacher feedback along student-generated rollouts rather than fixed teacher traces and has emerged as a promising post-training paradigm for long-horizon reasoning.
However, standard OPD typically generates full rollouts during training, which is computationally expensive and may expose the student to unreliable teacher feedback at late rollout positions, especially during early training.
We identify the rollout horizon as a key bottleneck in OPD that substantially impacts training efficiency. Unlike Reinforcement Learning with Verifiable Rewards (RLVR), OPD does not require a final answer reward to provide learning signals. Therefore, full rollouts may not always be necessary for OPD. Motivated by this insight, we propose two simple horizon-control strategies: \textbf{P}rogressive \textbf{OPD} (POPD), which gradually expands the rollout horizon during training, and \textbf{T}runcated \textbf{OPD} (TOPD), which permanently performs distillation on reliable truncated rollouts.
Experiments on mathematical reasoning show that POPD improves the training efficiency of OPD by up to 3$\times$, while TOPD matches OPD performance using only 10\% of the rollout horizon, leading to substantial wall-clock and memory reductions.
These results demonstrate that controlling the rollout horizon offers a simple and practical path to more efficient OPD.
\end{abstract}

\begin{figure*}[t]
    \centering
    \includegraphics[width=0.9\linewidth]{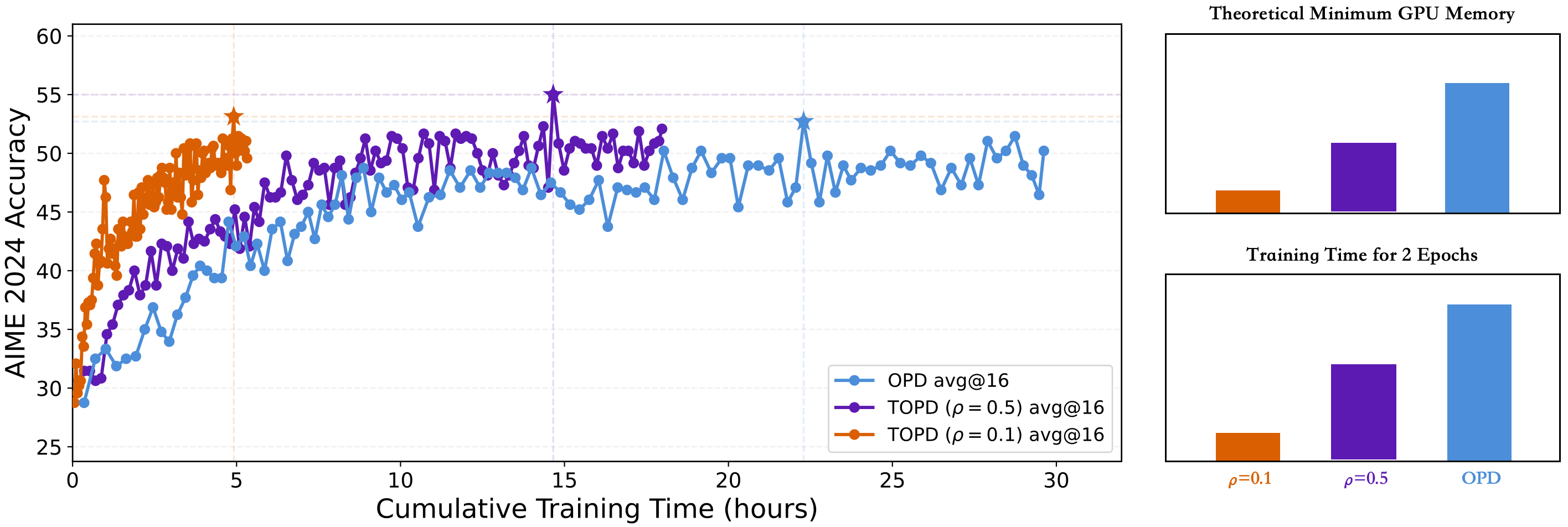}
    \caption{
\textbf{TOPD achieves comparable reasoning performance with substantially lower cost.}
Left: AIME24 accuracy curve for OPD and TOPD with different truncation ratios. Truncated variants, $\rho=0.1$ and $ \rho=0.5$, achieve comparable or better performance than full-rollout OPD, while requiring much less training cost. 
Right: Theoretical minimum GPU memory requirement and total training time for two epochs under different methods. 
Because TOPD generates only truncated rollouts and performs distillation on them, it substantially reduces wall-clock time and GPU memory footprint.
}
    \label{fig:topd_cumulative_time}
\end{figure*}

\section{Introduction}

Large language models (LLMs) have achieved remarkable progress on complex reasoning tasks, such as mathematical reasoning, code generation and question-answering \citep{shao2024deepseekmathpushinglimitsmathematical,feng2025retoolreinforcementlearningstrategic,zhang2025criticsearchfinegrainedcreditassignment}. On-policy distillation (OPD) \citep{agarwal2024onpolicy,gu2024minillm,yue2025does,lu2025onpolicydistillation}, which optimizes models on rollouts generated by the model itself rather than fixed teacher traces, has rapidly emerged as a core technique for LLM post-training. Compared with supervised fine-tuning (SFT) on fixed demonstrations, on-policy learning exposes the model to its own prefix distribution and directly improves behaviors that arise during generation. Recent industry efforts, including Qwen3 \citep{yang2025qwen3technicalreport}, Thinking Machines Lab \citep{lu2025onpolicydistillation}, MiMo \citep{coreteam2026mimov2flashtechnicalreport}, and GLM-5 \citep{glm5team2026glm5vibecodingagentic}, have incorporated OPD into their post-training pipelines, demonstrating its effectiveness for LLM reasoning.

Despite providing token-level supervision, OPD does not necessarily achieve high efficiency in long-horizon settings \citep{li2026rethinkingonpolicydistillationlarge}. 
Standard OPD requires the student to generate a full rollout and then computes log-ratio signals between teacher and student over the generated rollout. This design introduces two inefficiencies. 
First, generating full rollouts is computationally expensive, requiring substantial decoding time, KV cache memory, and log-probability computation. 
Second, the supervision quality of late rollout positions may be noisy (see Section~\ref{exp:disll_segment}), especially during early training. 
When the student policy differs substantially from the teacher policy, small deviations in early prefixes can accumulate over time, causing later tokens to drift away from regions where the teacher can provide reliable feedback.
Therefore, teacher feedback at late positions may become noisy, biased, or even misleading \citep{li2026rethinkingonpolicydistillationlarge,fu2026revisitingonpolicydistillationempirical}. This problem is further amplified in sequence-level OPD, where the update for an early token depends on a return-to-go style accumulation of future log-ratio signals. Thus, unreliable signals from late positions can be propagated backward and corrupt the gradients assigned to early tokens. 
This analysis suggests that OPD in long-horizon tasks suffers not only from computational cost, but also from degraded teacher reliability and the accumulation of future noise.

A natural remedy is token-level OPD, which aligns each generated token only with teacher feedback at the same position. By shortening the log-ratio horizon, token-level OPD avoids directly propagating noisy future signals to gradients at earlier positions, unlike sequence-level OPD. However, this alone is insufficient for efficient long-horizon training. Token-level OPD still generates full rollouts and distills all positions from the very beginning, even though it does not require return-to-go style accumulation of future log-ratio signals. As a result, the student spends substantial computation on late tokens whose supervision may be unreliable, especially in early training stages.
This observation reveals rollout horizon as another important dimension of OPD design. Therefore, efficient OPD should follow two principles: using a shorter log-ratio horizon to avoid the accumulation of future noise, 
and prioritizing distillation from reliable rollout segments to avoid premature learning from unreliable feedback at late rollout positions.

\begin{figure*}[t]
    \centering
    \includegraphics[width=1\linewidth]{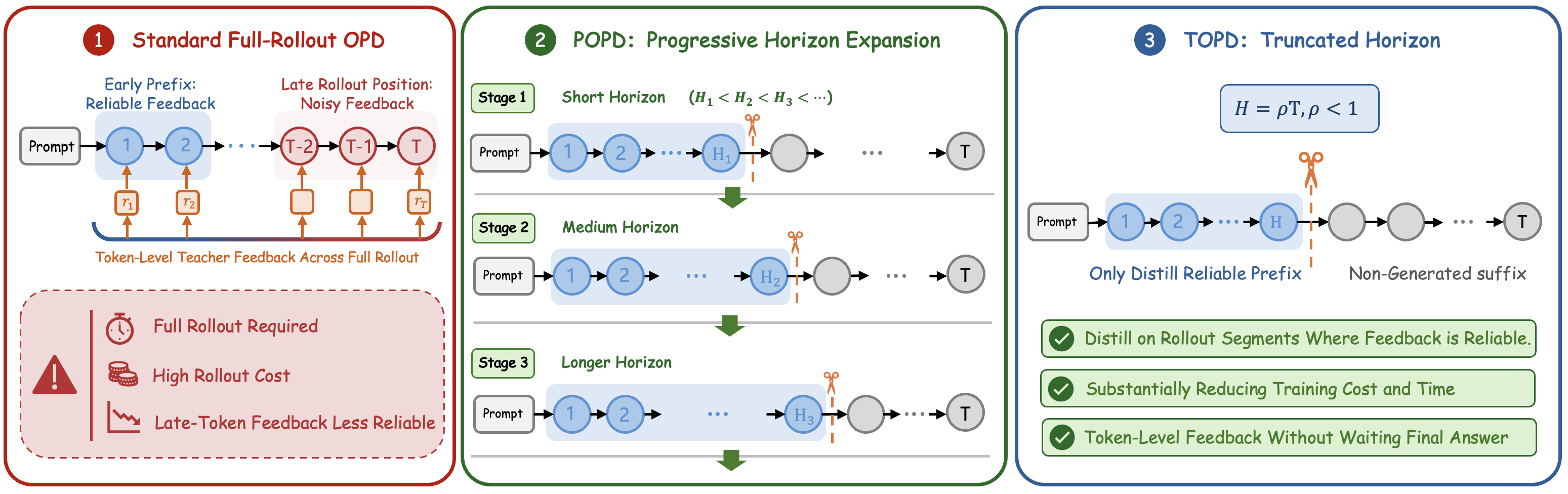}
    \caption{
\textbf{Overview of horizon control for efficient OPD.}
Standard OPD distills full rollouts throughout training. In contrast, POPD progressively expands the rollout horizon, while TOPD restricts distillation to truncated rollouts.
}
    \label{fig:framework.png}
\end{figure*}

Motivated by these principles, we propose two simple horizon-control strategies for efficient OPD (Fig.~\ref{fig:framework.png}). 
The first is \textbf{P}rogressive \textbf{OPD} (POPD), which starts training with a short rollout horizon and gradually expands the rollout horizon as training progresses. 
This curriculum allows the student to first align with the teacher on reliable prefixes before being exposed to longer reasoning trajectories. By avoiding noisy supervision on late rollout positions in the early stage, POPD improves training performance while reducing unnecessary generation cost. 
The second is \textbf{T}runcated \textbf{OPD} (TOPD), a simpler variant that permanently uses a truncated rollout horizon and distills only the reliable rollout segments. 
Unlike RLVR \citep{shao2024deepseekmathpushinglimitsmathematical,feng2025retoolreinforcementlearningstrategic}, token-level OPD does not require a complete rollout or wait for a final answer to provide learning signals. Therefore, even truncated rollout segments can provide meaningful supervision.

LLM reasoning experiments show that POPD improves the training efficiency of OPD and reaches strong performance with much less cumulative training time. TOPD further demonstrates that distilling from truncated but reliable rollout segments is sufficient to learn the teacher's reasoning patterns, without requiring teacher feedback over complete rollouts. In summary, our main contributions are as follows:
\vspace{-0.3em}
\begin{itemize}[leftmargin=5.0mm,label=$\circ$]
    \item We reveal rollout horizon control as a simple yet effective principle for efficient OPD in LLM reasoning tasks, and propose POPD and TOPD to improve training efficiency while achieving stronger distillation performance.
    \vspace{-0.5em}
    \item We provide detailed experimental analyses showing that distilling only partial rollout segments is sufficient to learn the teacher's reasoning patterns, and that earlier segments provide more effective supervision signals.
    \vspace{-0.5em}
    \item Extensive experiments show that POPD improves the training efficiency of OPD by up to 3$\times$. Meanwhile, TOPD can recover the training gains of standard OPD using only 10\% of the rollout horizon, reducing training time by up to 82\% in our experiments.
    \vspace{-0.5em}
\end{itemize}

\section{Preliminaries}

We consider on-policy distillation (OPD) for autoregressive language models.
Given an input prompt $x$, let $\pi_\theta$ denote the student policy and
$\pi^{\mathrm{g}}$ denote the teacher policy. The student generates a response
$y=(y_1,\ldots,y_T)$ from $\pi_\theta(\cdot\mid x)$. At each position $t$, we
define the log-ratio signal $r_t$ and the student score function $s_t$ as
\begin{equation}
\small
r_t =
\log
\frac{
\pi^{\mathrm{g}}(y_t \mid y_{<t}, x)
}{
\pi_\theta(y_t \mid y_{<t}, x)
},
\quad
s_t =
\nabla_\theta \log \pi_\theta(y_t \mid y_{<t}, x).
\label{eq:token_signal}
\end{equation}
Here, $r_t$ measures the difference between the teacher and student policies on the token generated by the student, while $s_t$ gives the gradient direction for increasing the
probability of that token.   

\paragraph{Sequence-Level OPD.}
To align with the teacher response, the OPD gradient can be written as
\begin{equation}
\small
\begin{aligned}
&\nabla_\theta J_{\mathrm{OPD}}(\theta)
= 
\mathbb{E}_{x,\,y\sim\pi_\theta} 
\Big[
\mathcal{L}(x,y) \nabla_\theta \log \pi_\theta(y\mid x)
\Big],
\end{aligned}
\label{eq:seq_opd}
\end{equation}
where $\small\mathcal{L}(x,y)=\big(
\log \pi^{\mathrm{g}}(y\mid x)-\log \pi_\theta(y\mid x)
\big)$. Using the autoregressive factorization, we have
\begin{equation}
\small
\mathcal{L}(x,y)
=
\sum_{k=1}^{T} r_k,
\quad
\nabla_\theta \log \pi_\theta(y\mid x)
=
\sum_{t=1}^{T} s_t.
\end{equation}
Since past log-ratio terms $r_k$ with $k<t$ do not depend on the current token
$y_t$, their expected contribution to $s_t$ is zero. Therefore, the sequence-level OPD gradient can be written as follows:
\begin{equation}
\small
\mathbb{E}\left[g^{\mathrm{seq}}\right]
=
\mathbb{E}_{x,\,y\sim\pi_\theta}
\left[
\sum_{t=1}^{T}
\left(
\sum_{k=t}^{T} r_k
\right)
s_t
\right].
\label{eq:seq_return_to_go}
\end{equation}

\paragraph{Discounted OPD.}
More generally, a discount factor $\gamma\in[0,1]$ can be applied to future OPD signals:
\begin{equation}
\small
\mathbb{E}\left[g^{\gamma}\right]
=
\mathbb{E}_{x,\,y\sim\pi_\theta}
\left[
\sum_{t=1}^{T}
\left(\sum_{k=t}^{T}
\gamma^{k-t} r_k\right) s_t
\right].
\label{eq:discounted_opd}
\end{equation}
When $\gamma$ is close to $1$, each token receives supervision from a long
future horizon. This introduces future information, but it may also
accumulate noisy or unreliable teacher signals from later positions.

\paragraph{Token-Level OPD.} Token-level OPD is the special case $\gamma=0$, where each token is only aligned with the teacher at the same position:
\begin{equation}
\small
\mathbb{E}\left[g^{\mathrm{tok}}\right]
=
\mathbb{E}_{x,\,y\sim\pi_\theta}
\left[
\sum_{t=1}^{T}
r_t s_t
\right].
\label{eq:token_opd}
\end{equation}
This approximate optimization formulation avoids directly propagating future log-ratio signals to earlier tokens, which motivates our analysis of OPD.

\section{Long-Horizon OPD Is Inefficient}
\subsection{Degraded Teacher Reliability}
\label{Degraded_Reliability}
\begin{figure}[t]
    \centering
    \includegraphics[width=1\linewidth]{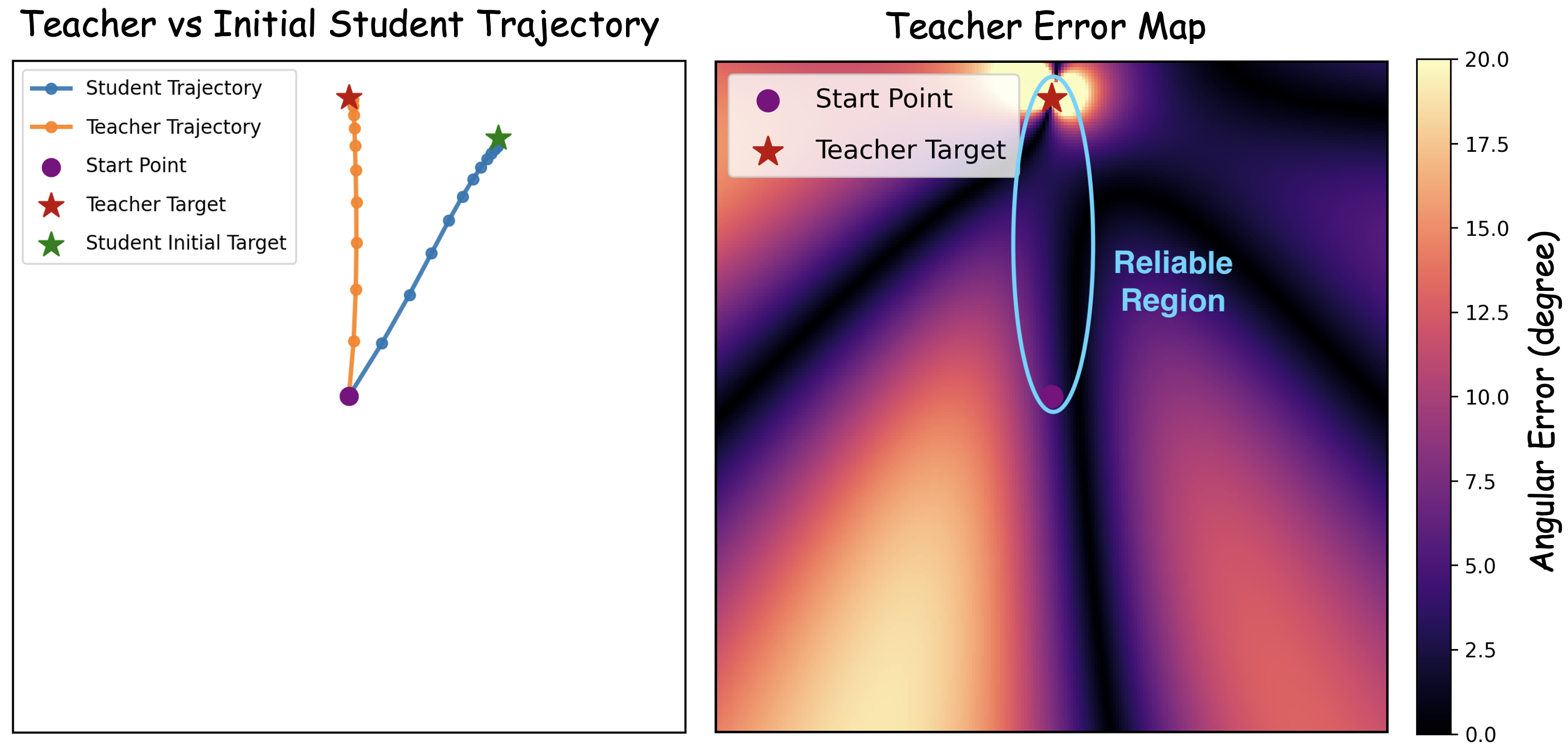}
    \caption{
\textbf{Teacher reliability in the simple navigation task.} The teacher and initial student trajectories point to different targets. Teacher guidance is reliable near its trajectory but becomes less reliable with distance.
}
    \label{fig:noise}
\end{figure}

\begin{promptbox}{Simple Navigation Task}
We consider a 2D control task to analyze how policy mismatches between teacher and student policies affect OPD (Fig.~\ref{fig:noise}). 
The teacher policy is trained with REINFORCE \citep{williams1992simple} toward a predefined teacher target, while the student initial policy is independently trained with REINFORCE toward a distinct student initial target. 

\vspace{0.5em}

At each time step, the state is represented as $s_t=[\mathrm{task\ id}, \mathrm{position}, t]$, where IDs 0 and 1 denote the student initial and teacher tasks, respectively.
The action $a_t=[\Delta x,\Delta y]$ specifies the movement along the two spatial dimensions, and the state transition follows
\[
s_{t+1}=s_t+a_t+z,\quad z\sim\mathcal{N}(0,\sigma).
\]
\end{promptbox}

As shown in Fig.~\ref{fig:noise}, to mimic teacher-student mismatch in LLM OPD, we use a 2D navigation task where the teacher and initial student are trained toward different targets. The angular difference between their targets controls the mismatch degree. Teacher feedback is reliable near its own high-probability trajectory region, but becomes increasingly noisy as states move farther away.
This indicates that teacher reliability is state-dependent rather than uniform over the full rollout space.

\begin{figure*}[t]
    \centering

     \includegraphics[width=1\linewidth]{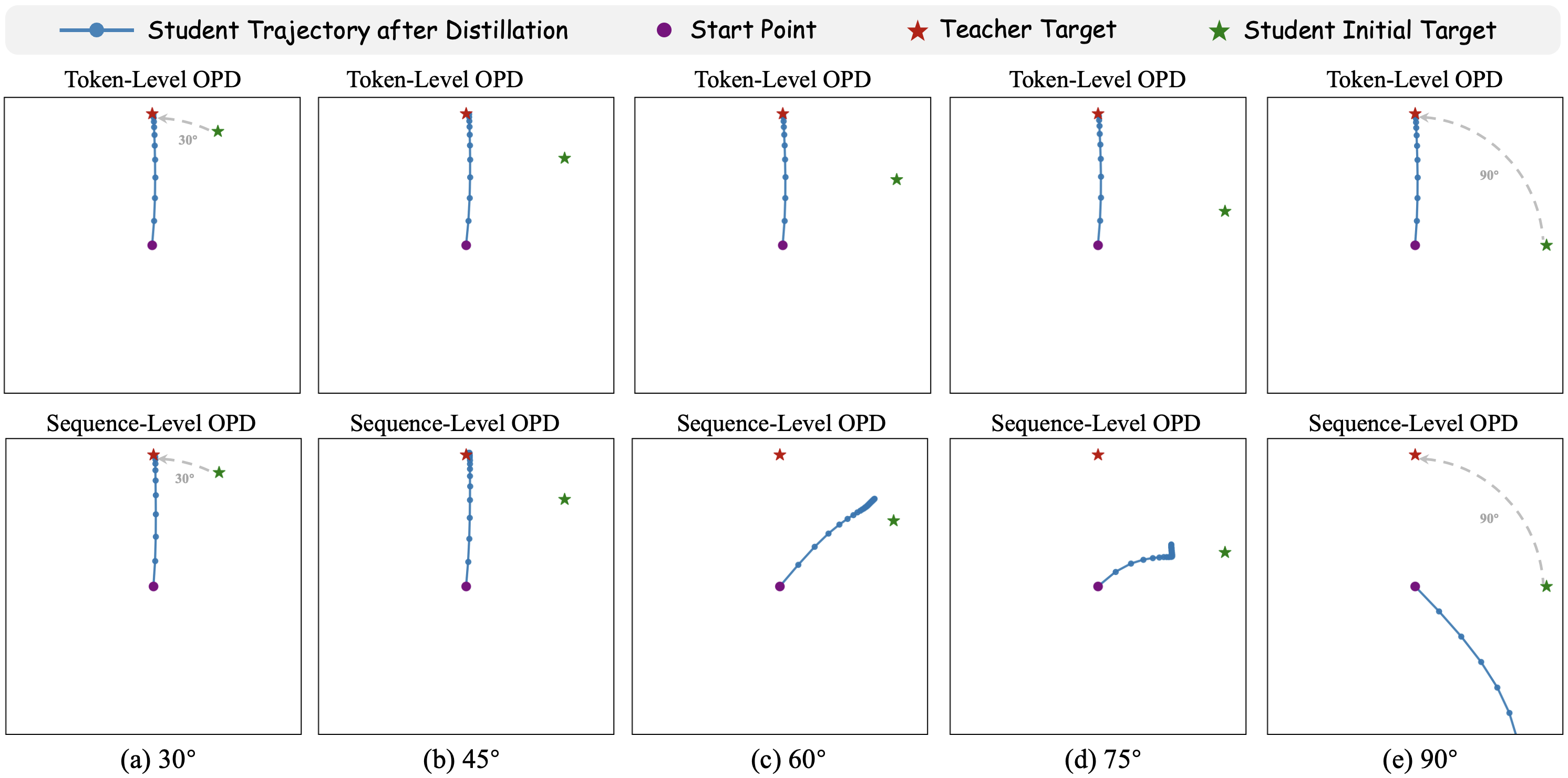}

\caption{
\textbf{Comparison between token-level OPD and sequence-level OPD under different degrees of teacher-student mismatch.}
When the mismatch is small ($30^\circ,45^\circ$), both methods can distill the student toward teacher Target.
As the mismatch increases ($60^\circ$, $75^\circ$, $90^\circ$), later rollout positions are more likely to enter regions with unreliable teacher feedback, causing sequence-level OPD to propagate noisy log-ratio signals back to early positions.
In contrast, token-level OPD remains stable by relying only on the log-ratio signal at the current position (Eq.~\eqref{eq:token_opd}).
}
    \label{fig:two_by_five}
\end{figure*}

\subsection{Accumulation of Future Noise in Sequence-Level OPD}

We next analyze why sequence-level OPD is inefficient under a mismatch between teacher and student policies. 
As shown in Eq.~(4), the update at position $t$ depends on the accumulated future log-ratio signals from positions $t$ to $T$. 
While such a return-to-go formulation can exploit future information, it makes early tokens sensitive to unreliable teacher signals that appear in later rollout position.

Fig.~\ref{fig:two_by_five} illustrates this effect in the navigation task. 
When the mismatch between the teacher and student (the angular difference between teacher and student initial targets) is small, student rollouts remain close to the trajectory distribution familiar to the teacher, and both token-level and sequence-level OPD provide effective guidance. 
However, under a larger mismatch, later rollout positions are more likely to enter regions where teacher feedback is noisy or biased. 
In this case, sequence-level OPD propagates these unreliable log-ratio signals back to earlier tokens during training, whereas token-level OPD effectively avoids this propagation of noise by only relying on the current log-ratio signal. We formalize this intuition with a simple noise model.  The derivation is provided in Appendix~\ref{app:future_noise_derivation}.

\begin{proposition}[Accumulation of Future Noise]
\label{prop:future_noise}
Consider the sequence-level OPD advantage at position $t$:
\begin{equation}
\small
    \hat{A}^{\mathrm{seq}}_t
    =
    \sum_{k=t}^{T} \gamma^{k-t} r_k,
\end{equation}
where $r_k$ is the log-ratio signal at position $k$. Suppose that $r_k$ can be decomposed as
\begin{equation}
\small
    r_k = r_k^\star+ \sigma_k z_k,
    \qquad
    z_k \sim \mathcal{N}(0,1),
\end{equation}
where $r_k^\star$ is the ideal log-ratio signal and $\sigma_k z_k$ denotes stochastic noise. 
In particular, when $\gamma=1$ and the noise scale grows with the rollout position, e.g., $\sigma_k=\delta k$, the mean-squared error of the sequence-level advantage satisfies
\begin{equation}
\small
    \mathrm{MSE}_t 
    \left(
    \hat{A}^{\mathrm{seq}}_t - A^{\star}_t
    \right)
    =
    \delta^2
    \sum_{k=t}^{T} k^2.
\end{equation}
Thus, earlier tokens accumulate more future noise, with error growing as $O(T^3)$ when $t=1$.
\end{proposition}

When the teacher is optimal, the noise term $\sigma_k z_k$ in Proposition \ref{prop:future_noise} vanishes, so future log-ratio signals can still provide accurate guidance. Therefore, sequence-level OPD can successfully distill the teacher policy under an optimal teacher compared to a suboptimal teacher (Appendix~\ref{app:Optimal_Teacher_Policy_simple_task}). However, in practical LLM post-training scenarios, the teacher policy is rarely optimal.

\subsection{Design Principles}

The above analysis suggests that efficient OPD should follow two design principles.

\paragraph{Principle 1: Shorten the log-ratio horizon.}
Sequence-level OPD assigns each token a return-to-go style signal that accumulates future log-ratio terms between teacher and student. 
As shown in Proposition~\ref{prop:future_noise}, this accumulation can introduce noise from unreliable future positions, thereby corrupting the gradients of early tokens.
A natural way to avoid this issue is to use token-level OPD. As shown in Eq.~\eqref{eq:token_opd}, instead of assigning a cumulative future signal to each token, token-level OPD only aligns the student with the teacher feedback at the same position.
This form prevents unreliable future log-ratio signals from being propagated to earlier tokens, thereby leading to more efficient training. 

\paragraph{Principle 2: Control the rollout horizon.}
A shorter log-ratio horizon alone is not sufficient for efficient OPD. Even with token-level OPD, standard OPD still generates and distills full rollouts from the beginning of training. Distilling tokens at these late positions can increase training cost while providing only limited or noisy learning signals (section \ref{exp:disll_segment}). 
Moreover, unlike sequence-level OPD, token-level OPD does not require generating full rollouts to compute the return-to-go style log-ratio signal. Therefore, efficient OPD should also avoid generating full rollouts too early or when they are unnecessary.

\section{Methods}

\subsection{Progressive On-Policy Distillation}

In this section, we first introduce \textit{Progressive OPD} (POPD), a simple horizon curriculum for improving the efficiency of OPD in long-horizon tasks. 
The key idea of POPD is to start distillation from short rollout segments and gradually expand the rollout horizon as training progresses. This allows the student to first align with the teacher on reliable early prefixes, before being exposed to longer reasoning trajectories.
Instead of using the full trajectory from the beginning, POPD only applies OPD to the first $H_k$ positions at training step $k$:
\begin{equation}
\small
    \mathbb{E}\left[g_{\mathrm{popd}}\right]
    =
    \mathbb{E}_{x, y \sim \pi_\theta}
    \left[
    \sum_{t=1}^{\textcolor{blue}{H_k}} r_t s_t
    \right],
\end{equation}
where $H_k$ is the rollout horizon used at training step $k$. We use a progressive schedule:
\begin{equation}
\small
    H_k =
    \min\left(
    T,\,
    H_0 +  \Delta H \left\lfloor\frac{k}{\Delta k} \right\rfloor
    \right),
\label{eq:popd_horizion}
\end{equation}
where $H_0$ is the initial rollout horizon, $\Delta H$ is the horizon increment, and ${\Delta k}$ is the interval for increasing the horizon. Here, $\lfloor \cdot \rfloor$ denotes the floor operation. Thus, the rollout horizon remains short at the beginning of training and is gradually increased after every ${\Delta k}$ training steps. 
This schedule reduces exposure to unreliable feedback at late rollout positions during early training and lowers generation cost.

\subsection{Truncated On-Policy Distillation}

As discussed in the previous section, a distinctive advantage of OPD over RLVR is that it does not require a complete trajectory to obtain the learning signal. Motivated by this observation, we consider another simple but effective variant, termed \textit{Truncated OPD} (TOPD). Instead of generating a full rollout of length $T$, TOPD only generates and distills the first $H$ tokens, where $H<T$. The token-level OPD gradient can be written as:
\begin{equation}
\small
    \mathbb{E}\left[g_{\mathrm{topd}}\right]
    =
    \mathbb{E}_{x, y \sim \pi_\theta}
    \left[
    \sum_{t=1}^{\textcolor{blue}{H}} r_t s_t
    \right].
\end{equation}
Compared with standard OPD, which uses all positions from $1$ to $T$, TOPD uses only the prefix of length $H$. When $H = \rho T_{\max}$, where $T_{\max}$ is the maximum response length and $\rho < 1$, the cost of the rollout generation and training is reduced approximately in proportion to $\rho$,
making long-horizon OPD substantially more affordable.

\section{Experiments}
\subsection{LLM Reasoning Experiments}

\begin{table}[h]
\centering
\caption{\textbf{The main results.} \textbf{Bold} values indicate the best result. The accuracy is reported as the average of avg@16 over the last 10 evaluations during training. The reported training times were measured on a node equipped with 8 NVIDIA H20 GPUs}
\resizebox{1\linewidth}{!}{
\begin{tabular}{@{}lcccccc}
\toprule
 \textbf{Method} & $\rho$ & \textbf{AIME24}    & \textbf{AIME25} & \textbf{AMC23} & \textbf{Avg} & \textbf{Time$\downarrow$} \\ 
\midrule
\rowcolor{lightorange}\multicolumn{7}{l}{\textbf{Student}:  \textit{R1-Distill-1.5B}  \quad \textbf{Teacher}: \textit{JustRL-R1-1.5B}} \\
\midrule
OPD &N/A      & 49.3           & 34.3 & \textbf{80.7}  & 54.8 & 29.7h  \\
POPD &N/A      & 51.0           & \textbf{34.8} & 79.3  & \textbf{55.1} &10.1h  \\
TOPD  &0.50            & 50.3           & \textbf{35.1} & 80.0 & \textbf{55.1} &18.1h \\
TOPD &0.25      & \textbf{51.3}  & 34.2 & 79.5 & 55.0&10.2h\\
TOPD &0.10      & 50.7           & 32.5 & 76.6  & 53.3 &\textbf{5.3}h  \\
\midrule
\rowcolor{lightorange}\multicolumn{7}{l}{\textbf{Student}:  \textit{OpenMath-1.5B}  \quad \textbf{Teacher}: \textit{JustRL-Nemotron-1.5B}} \\
\midrule
OPD  &N/A    &       68.7 &  60.3 &  89.9 & 73.0&   39.6h\\
POPD &N/A    &       70.0 &  60.7 &  89.7 & 73.5&   10.0h\\
TOPD &0.50   &       70.0 &  61.0 &  90.2 & 73.7&   20.0h\\
TOPD &0.25   &       70.1 &  60.9 &  \textbf{90.6} & 73.9&   10.4h\\
TOPD &0.10   &       \textbf{70.7} &  \textbf{61.4} &  90.2 & \textbf{74.1}&   \textbf{5.2h} \\

\bottomrule

\label{tab:question_difficulty}
\end{tabular}
}
\end{table}

\paragraph{Setup.} 
{We use two student-teacher pairs in our LLM reasoning experiments. 
The first pair uses R1-Distill-1.5B \citep{deepseekai2026deepseekr1incentivizingreasoningcapability} as the student and JustRL-R1-1.5B \citep{he2025justrlscaling15bllm} as the teacher. 
The second pair uses OpenMath-1.5B \citep{du2025nemotronmathefficientlongcontextdistillation} as the student and JustRL-Nemotron-1.5B \citep{he2025justrlscaling15bllm} as the teacher. }
All student models are trained for 2 epochs on the DAPO-Math-17K dataset \citep{yu2025dapoopensourcellmreinforcement} and evaluated every 5 steps on AMC23, AIME24, and AIME25 using avg@16. 
During evaluation, all models are allowed to generate full rollouts. 
We analyze and compare three OPD variants: OPD\footnote{Unless otherwise specified, OPD refers to token-level OPD in the following experiments.}, TOPD, and POPD.

\begin{figure}[t]
    \centering
    \includegraphics[width=1\linewidth]{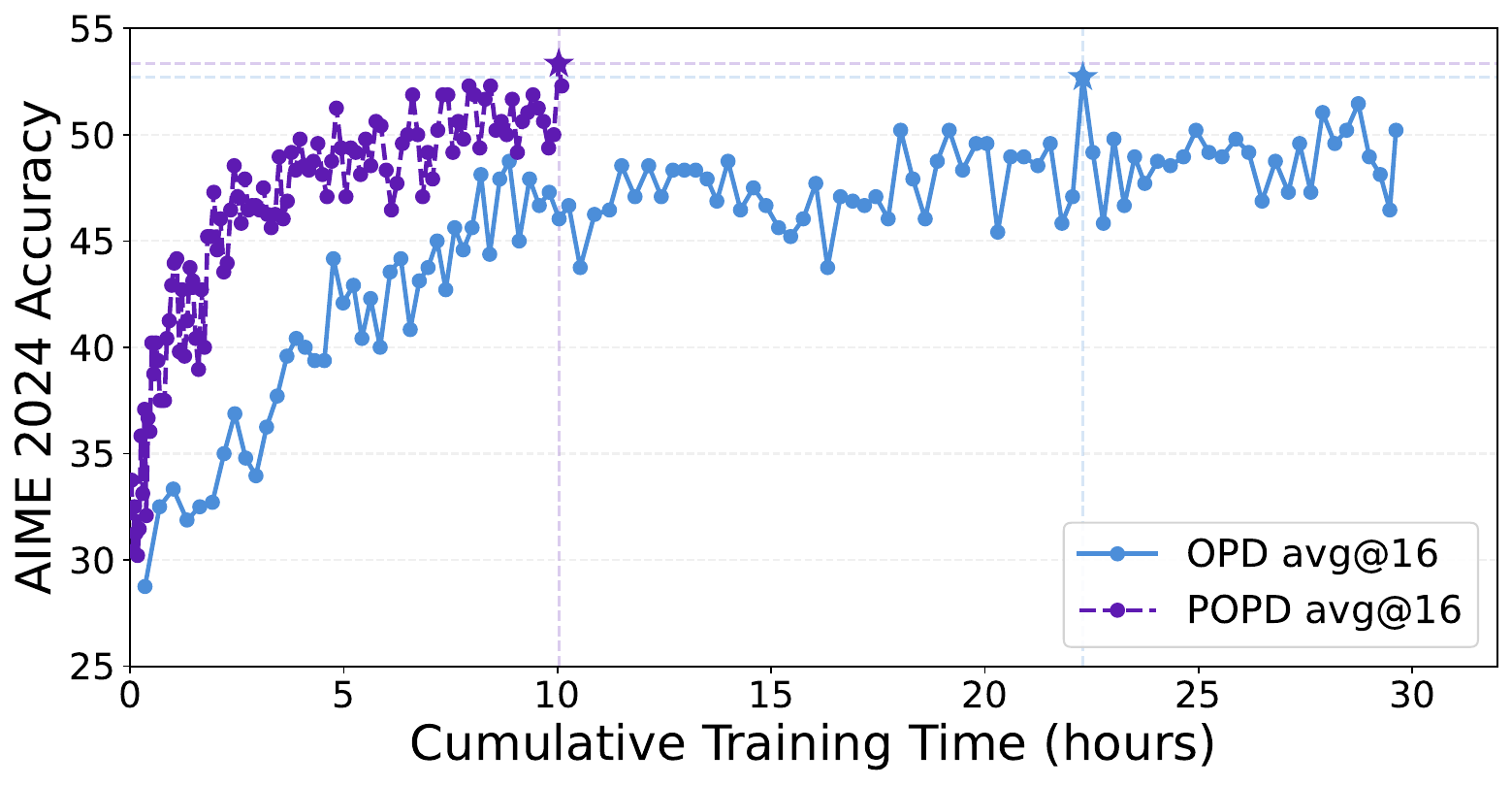}
    \caption{
\textbf{Training efficiency comparison between OPD and POPD.} The student is R1-Distill-1.5B, and the teacher is JustRL-R1-1.5B.
}
    \label{fig:popd_opd}
\end{figure}

\paragraph{Progressive OPD improves training efficiency.}
Fig.~\ref{fig:popd_opd} compares the training dynamics of standard OPD and POPD on AIME24. 
POPD reaches strong performance much faster than standard OPD. 
This result supports our hypothesis that standard OPD is inefficient in long-horizon reasoning tasks. Early in training, generating and distilling complete trajectories can waste computation on late positions where teacher feedback may be less reliable (see Section~\ref{exp:disll_segment}). By first training on shorter rollouts and gradually expanding the horizon, POPD reduces unnecessary long-horizon generation while preserving useful token-level supervision.

\begin{figure}[h]
    \centering
    \includegraphics[width=1\linewidth]{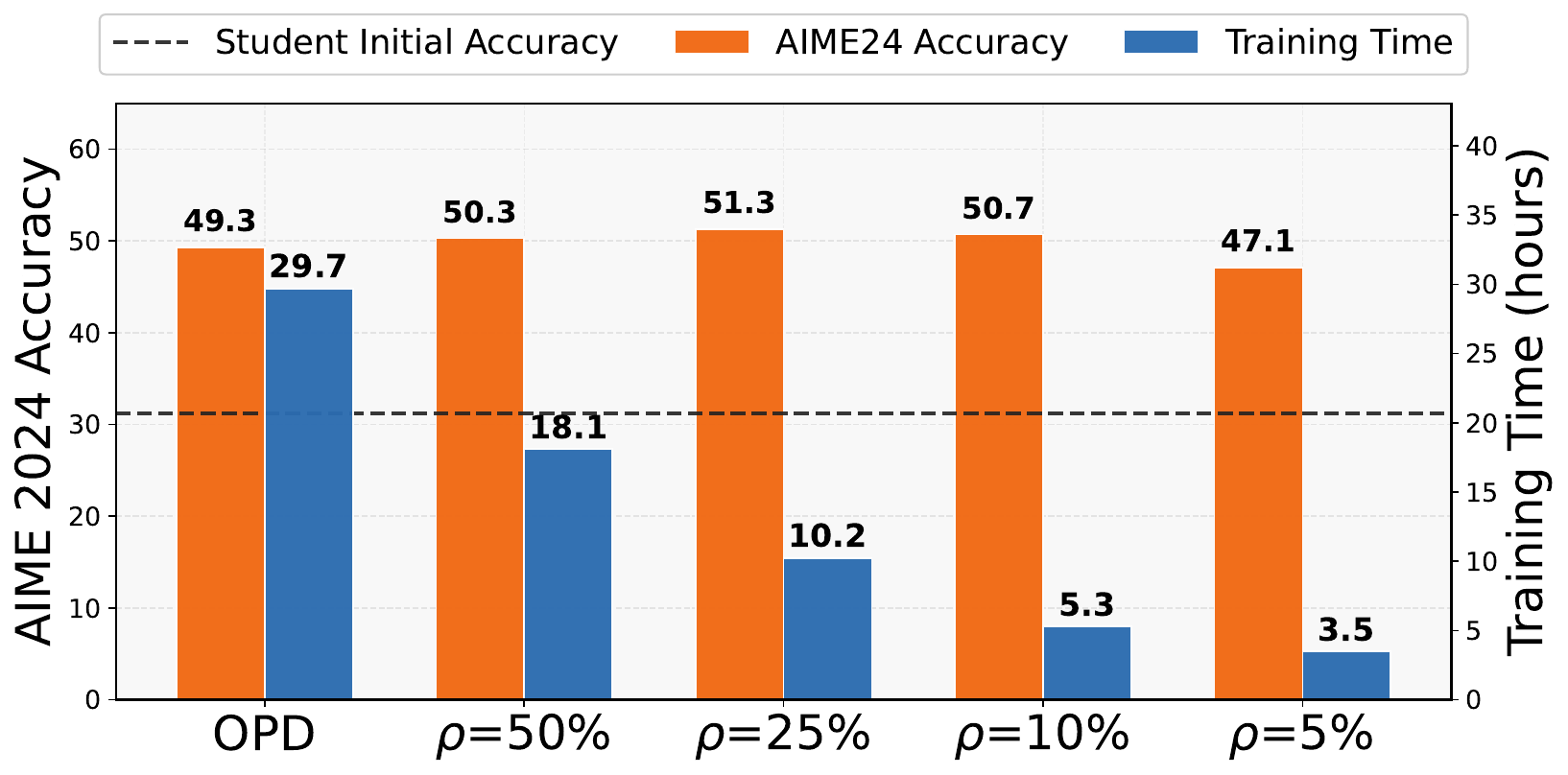}
    \caption{
\textbf{Performance of TOPD with different rollout ratios on AIME24.} The student is R1-Distill-1.5B, and the teacher is JustRL-R1-1.5B.
Moderate truncation matches or even surpasses the performance of standard OPD while substantially reducing training cost.}
    \label{fig:topd_compare}
\end{figure}

\paragraph{Truncated OPD provides a trade-off between cost and performance.}
Table ~\ref{tab:question_difficulty} and Fig.~\ref{fig:topd_compare} show that moderate truncation matches or outperforms standard OPD while substantially reducing training time. 
For example, TOPD with $\rho=25\%$ improves AIME24 accuracy from 49.3 to 51.3 and reduces training time from 29.7h to 10.2h. 
Even with $\rho=10\%$, TOPD retains most of the performance gain with only 5.3h of training.
These results show that OPD can effectively benefit from partial rollouts during training, since token-level log-ratio feedback does not require a final answer.

\begin{figure}[t]
    \centering

    \includegraphics[width=0.7\linewidth]{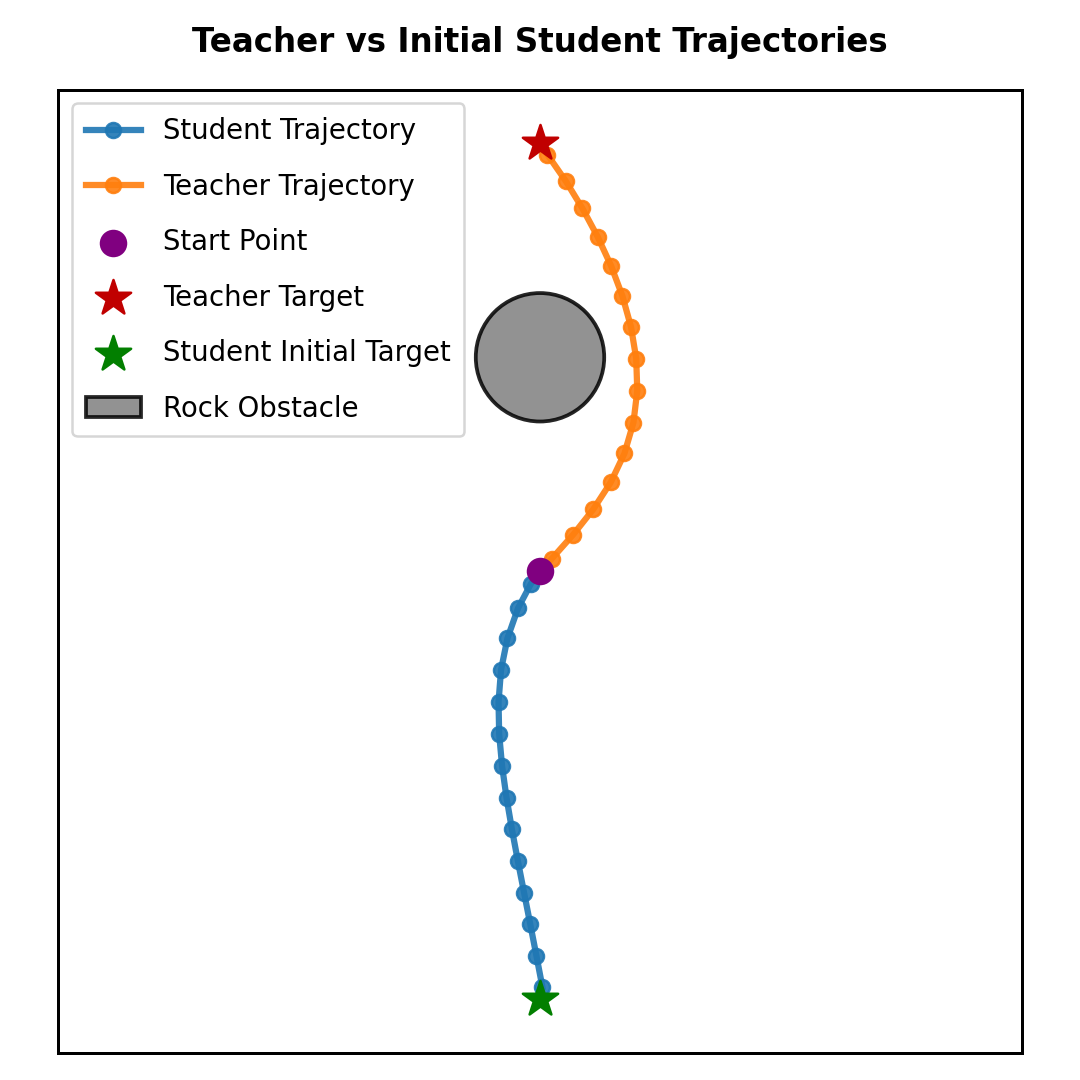}

    \caption{
\textbf{Autoregressive control task. } The teacher and initial student trajectories point to different targets. We need to distill teacher policy into student so that the trajectories generated by student head toward teacher target. Please refer to Appendix~\ref{app:autoregressive_navigation_task} for task details.}
\label{fig:auto_regressive_traj}
\end{figure}

\subsection{Analysis on Autoregressive Control Task}
\paragraph{Setup.} To further evaluate POPD and TOPD in a broader range of task scenarios, we introduce an autoregressive control task (see Fig.~\ref{fig:auto_regressive_traj}). Compared with the simple navigation task in Sec~\ref{Degraded_Reliability}, this environment requires the policy to make decisions based on historical velocity sequences, making the task more challenging.

\begin{figure}[h]
    \centering
    \includegraphics[width=0.97\linewidth]{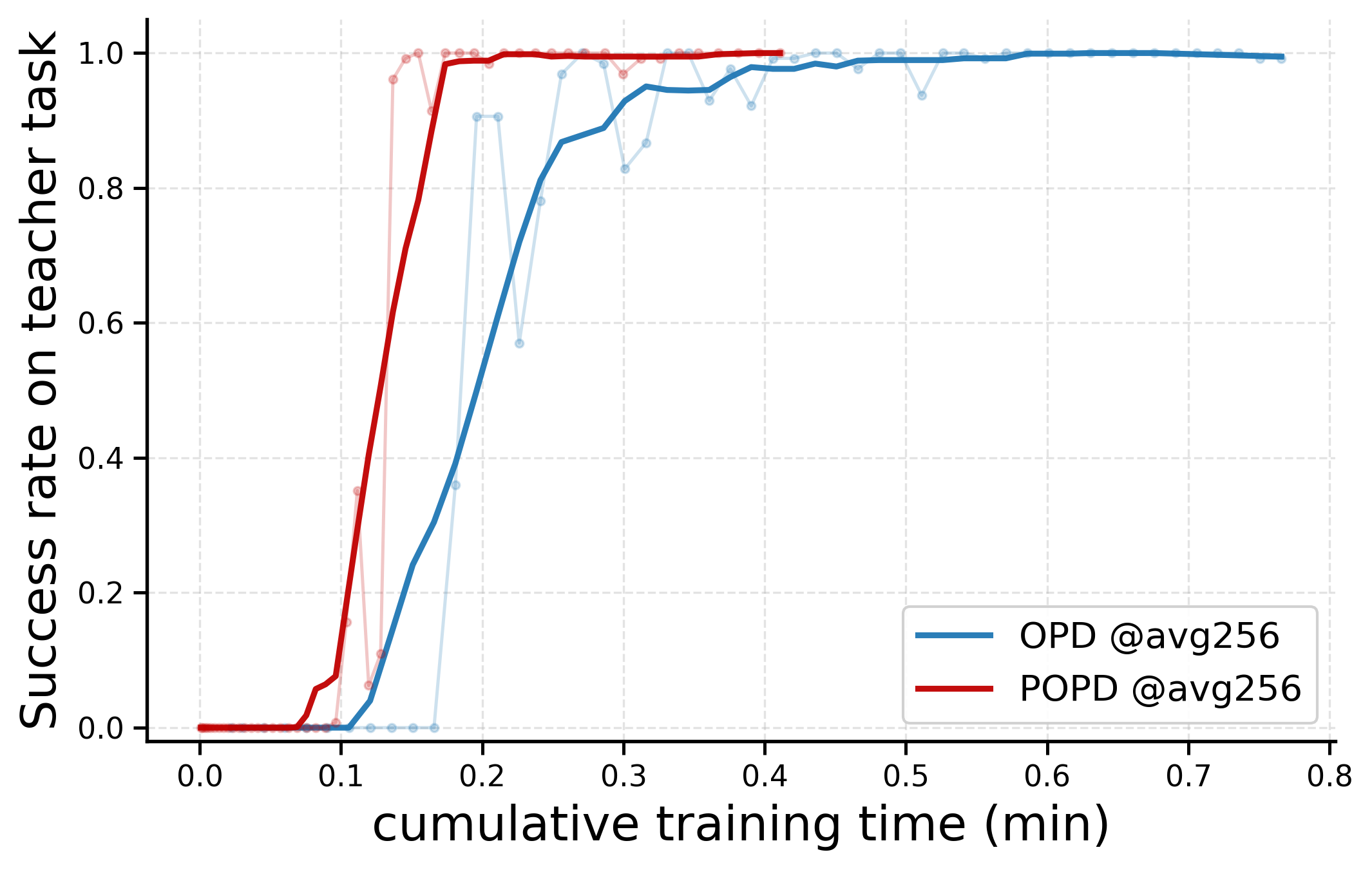}
    \caption{
\textbf{Training efficiency comparison between OPD and POPD in the autoregressive control task.} 
POPD reaches a high success rate substantially faster than standard OPD, showing that progressive horizon expansion improves long-horizon distillation efficiency.
}
\label{fig:auto_regressive_efficiency}
\end{figure}

\paragraph{POPD improves training efficiency again. } As shown in Fig.~\ref{fig:auto_regressive_efficiency}, POPD achieves a high success rate much faster than standard OPD. This again supports our claim that full rollouts can be inefficient, especially during early training, because late rollout positions may already deviate from the distribution familiar to the teacher and provide less reliable supervision. By starting from truncated rollouts and progressively expanding the horizon, POPD avoids premature learning from noisy feedback signals and accelerates convergence.

\begin{table}[h]
\centering
\caption{\textbf{Success rates of different methods under the autoregressive control task.} The reported success rate is taken from the best evaluation step.}
\resizebox{0.9\linewidth}{!}{
\begin{tabular}{ccc}
\toprule
 \textbf{\quad Method \quad }  & $\rho$ & \textbf{\quad Success Rate@avg256\quad} \\ 
\midrule
OPD &N/A      &  100\% \\
POPD &N/A     &  100\% \\
TOPD &0.25     & 93\%  \\
TOPD &0.20     & 0\%  \\
TOPD &0.10     & 0\%  \\
\bottomrule

\label{tab:topd_control_success_rate}
\end{tabular}
}
\end{table}

\paragraph{TOPD fails in the autoregressive control task.}
Unlike LLM reasoning, where early rollout segments often contain generalizable reasoning patterns, control tasks exhibit strong state-action coupling. Therefore, truncating distillation to the first $\rho$ portion of the rollout removes supervision from many critical states. In our navigation environment, this makes TOPD less effective (see Table~\ref{tab:topd_control_success_rate}), since the student may learn the initial moving direction but fail to acquire the later behaviors required to bypass the obstacle and reach the target. This suggests that TOPD is effective when useful teacher characteristics are independent of the horizon, but may fail in tasks where the teacher exhibits phase-dependent behaviors across different rollout stages.

\paragraph{Conclusion. } In summary, we can draw the following preliminary conclusions:
\begin{idea}{}
\begin{itemize}[leftmargin=4.0mm,label=$\circ$]
    \item POPD is effective across a variety of scenarios because it can eventually access the full rollout.
    \item TOPD is applicable in scenarios where the teacher's characteristics that we aim to learn are independent of the horizon and can be contained in a truncated rollout (e.g., reasoning patterns).
\end{itemize}
\end{idea}
\noindent In the next section, we further analyze why TOPD is effective in LLM reasoning tasks.

\section{Why Training on Truncated Rollouts Is Effective?}
\subsection{Prefix-continuation Analysis }

\begin{figure}[t]
    \centering
    \includegraphics[width=1\linewidth]{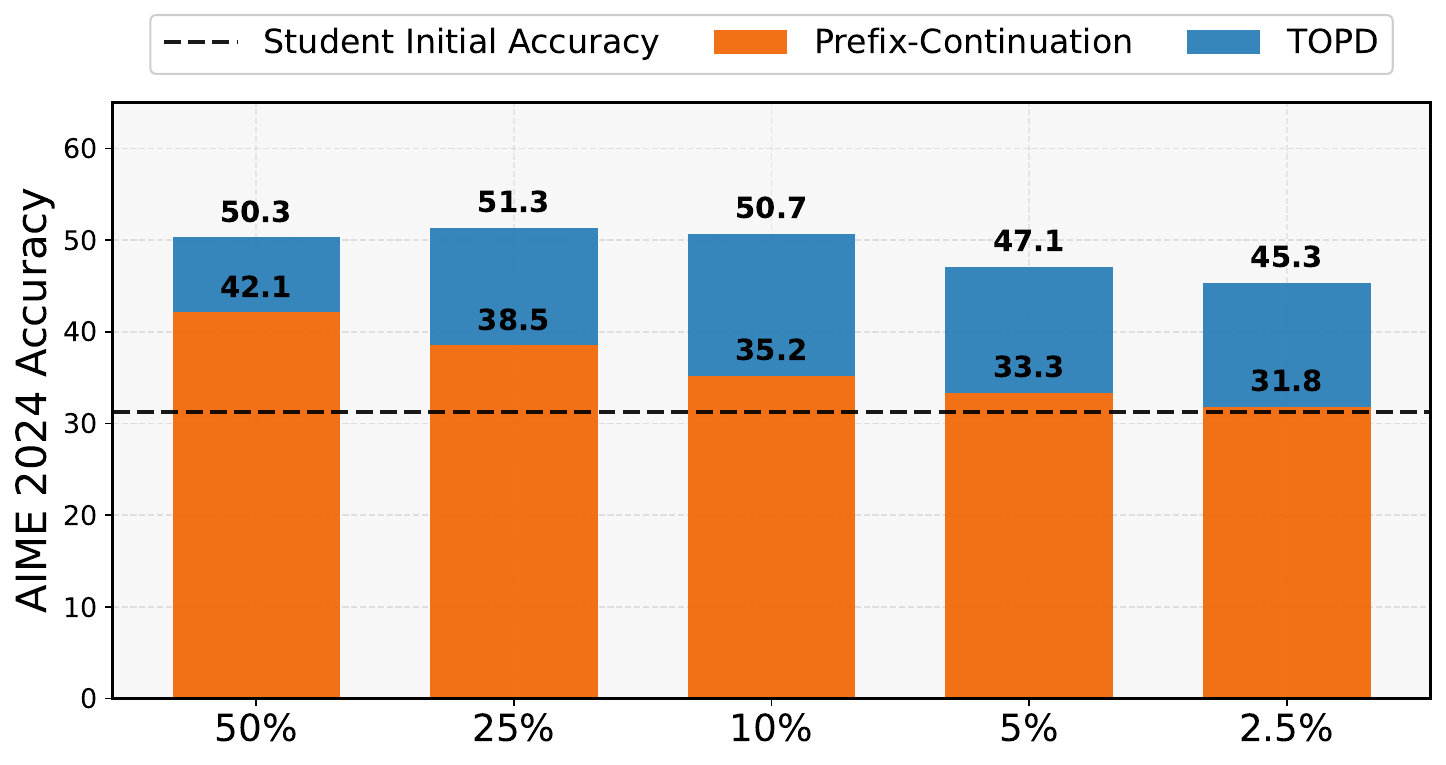}
    \caption{ \textbf{Prefix-continuation analysis on AIME24.} For each problem in AIME24, the teacher generates 256 truncated prefixes, and the student (without any OPD training) continues generation from each teacher-generated prefix. The x-axis denotes the rollout ratio of teacher-generated prefixes in the prefix-continuation setting, or the truncation ratio used in the TOPD setting.}
    \label{fig:topd_performance_prefix}
\end{figure}

The previous results show that OPD does not necessarily require full rollouts to improve the student model in LLM reasoning. 
Even when distillation is applied only to truncated rollouts, the student can still recover a large fraction of the performance of the teacher. We next examine whether TOPD works merely because teacher-generated prefixes activate better sampling paths. To isolate this prefix effect from parameter updating, we first conduct a prefix-continuation experiment without updating the student. The student is R1-Distill-1.5B, and teacher is JustRL-R1-1.5B. As shown in Fig.~\ref{fig:topd_performance_prefix}, continuing from teacher-generated prefixes with length ratio $\rho$ yields only modest improvement, far smaller than TOPD training on rollouts of $\rho$.  Therefore, 
\begin{idea}{}
The effectiveness of TOPD does not arise from teacher-generated prefixes activating successful sampling paths.
\end{idea}

\begin{figure}[t]
    \centering
    \includegraphics[width=1\linewidth]{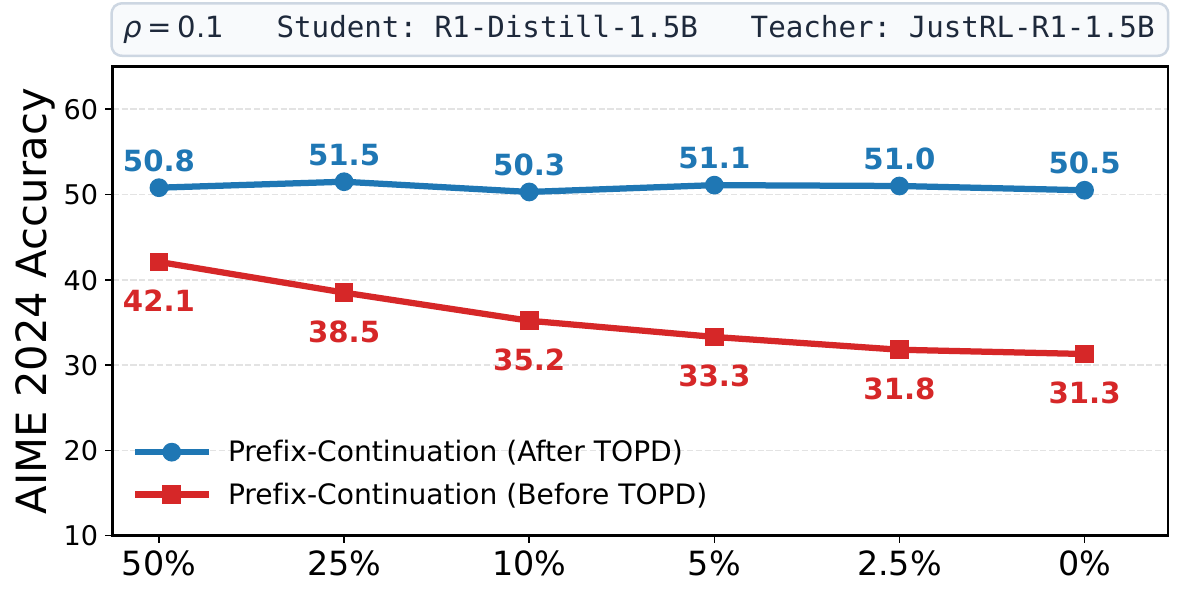}
    \caption{\textbf{Prefix-continuation analysis before and after TOPD.} We respectively use the student model before and after TOPD ($\rho=0.1$) training to continue generation from teacher-generated prefixes. The x-axis is the rollout ratio of the teacher-generated prefixes.}
    \label{fig:topd_performance_prefix_trained}
\end{figure} 

We then distill the student model using TOPD ($\rho = 0.1$) and again evaluate the distilled student by conditioning its generation on teacher-generated prefixes. This allows us to examine whether teacher-generated prefixes still influence a student that has already been trained with TOPD. As shown in Fig.~\ref{fig:topd_performance_prefix_trained}, the distilled student model achieves the same performance with or without teacher-generated prefixes, regardless of the length of those prefixes.
This indicates that:
\begin{idea}{}
The student has learned the teacher's reasoning pattern from rollout segments. Consequently, regardless of the length of the teacher-generated prefixes, all such prefixes appear as in-distribution rollouts to the student and thus do not affect its performance.
\end{idea}
Therefore, TOPD is effective not because imitating teacher prefixes can activate particular successful sampling paths, but because applying distillation on truncated rollouts truly changes the student's policy distribution, allowing it to acquire the teacher's horizon‑independent, generalizable reasoning patterns.

\subsection{Analysis of Reverse Distillation}
To further test whether TOPD provides a strong optimization signal, we conduct a reverse distillation experiment.
Here, the weaker R1-Distill-1.5B is used as the teacher, while the stronger JustRL-R1-1.5B is used as the student.
If OPD on truncated rollouts is indeed effective, then even when the teacher is weaker, the student should be pulled toward the teacher's behavior.

\begin{figure}[h]
    \centering
    \includegraphics[width=1\linewidth]{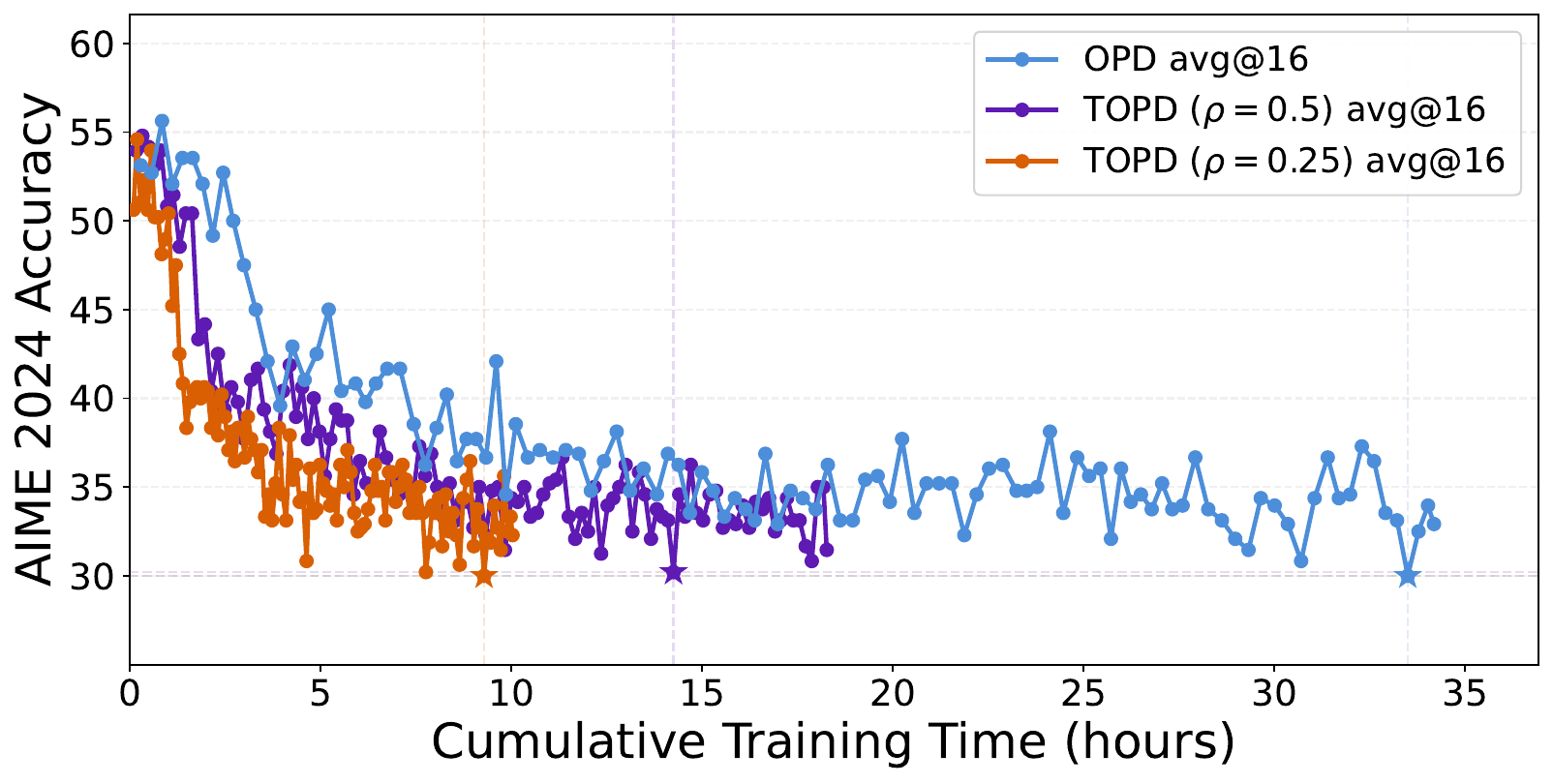}
    \caption{\textbf{Reverse distillation on AIME24.} We use the weaker R1-Distill-1.5B as the teacher and the stronger JustRL-R1-1.5B as the student. TOPD pulls the student toward the weaker teacher even with truncated rollouts, indicating that truncated rollouts provide a strong optimization signal.}
    \label{fig:reverse_distill}
\end{figure}

As shown in Fig.~\ref{fig:reverse_distill}, reverse distillation decreases the performance of JustRL-R1-1.5B toward the level of R1-Distill-1.5B, even when training over truncated rollouts. This also confirms that on-policy distilling on truncated rollouts is strong enough to reshape the student policy, without full rollouts or final answer supervision.
Together with the prefix-continuation analysis, this result supports our claim that TOPD works through genuine parameter updating. Truncated rollouts already contain sufficient information about the teacher's reasoning patterns, including problem decomposition, intermediate setup, and initial solution direction. By aligning the student with these early rollout segments, OPD can efficiently transfer the teacher's reasoning patterns to the student.

\subsection{Where Do Useful Distillation Signals Come From?}
\label{exp:disll_segment}
To understand why TOPD can match or even outperform standard OPD, we conduct a segment-level distillation analysis. 
We split each rollout into several 10\% segments, including 0--10\%, 20--30\%, ..., and 80--90\%, and distill the student using only one segment at a time while masking out the remaining tokens. This setup investigates whether arbitrary rollout segments can provide sufficient supervision for learning the teacher's reasoning patterns.

\begin{figure}[h]
    \centering
    \includegraphics[width=1\linewidth]{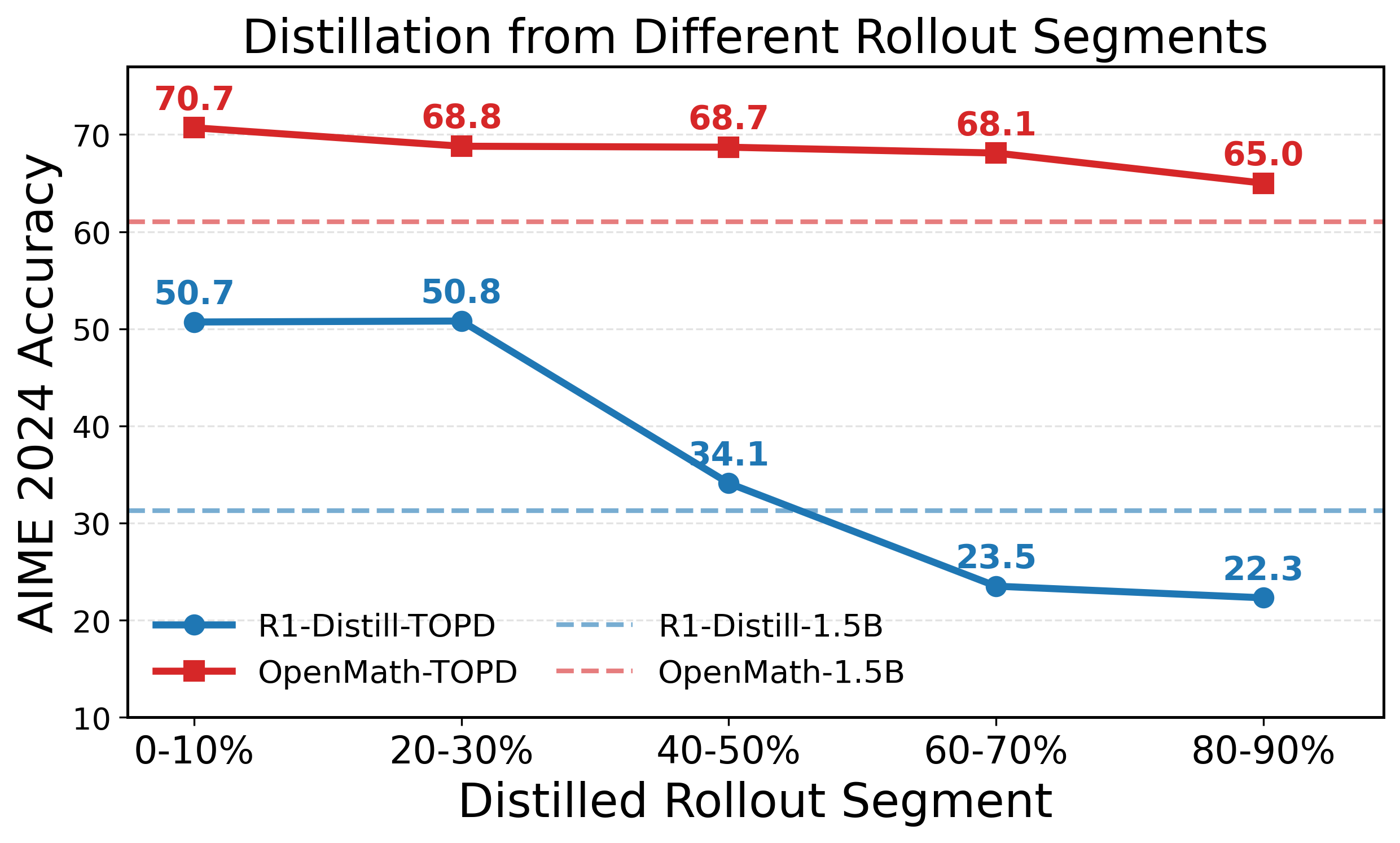}
    \caption{\textbf{Ablation study on distillation from different rollout segments.} We split each complete rollout into several segments according to token positions and distill the student using one segment at a time.}
    \label{fig:different_rollout_segment}
\end{figure}

\begin{figure}[h]
    \centering
    \includegraphics[width=1\linewidth]{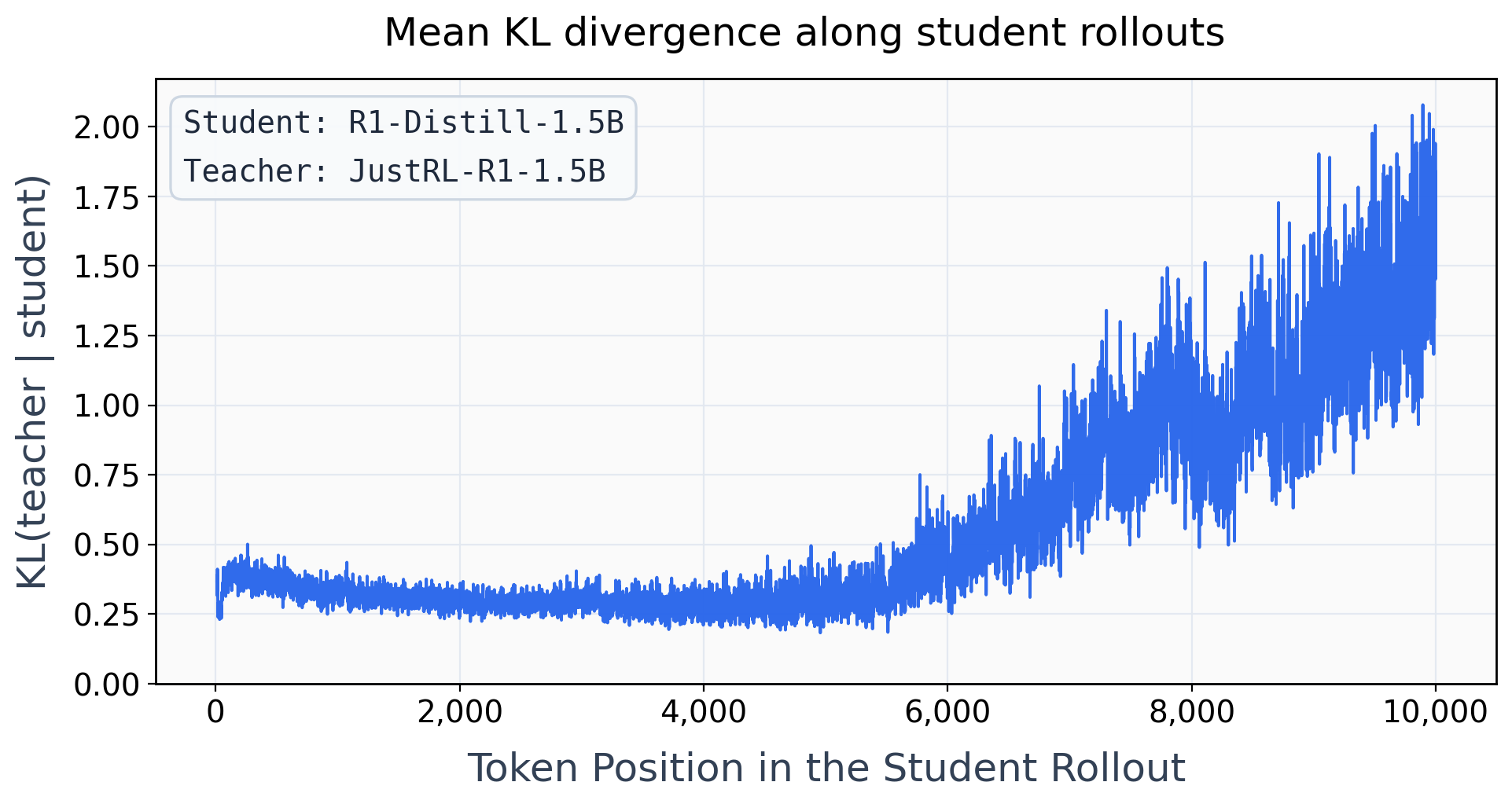}
    \caption{We randomly sampled 500 questions from DAPO-Math-17K, and measured the KL divergence between the teacher and the student at each position within the student rollout. It can be observed that as the rollout position moves deeper, the KL divergence increases, indicating that the end of student's rollout gradually shifts to a distribution unfamiliar to the teacher.}
    \label{fig:position_kl_curve}
\end{figure}

As shown in Figure~\ref{fig:different_rollout_segment}, early and middle rollout segments provide more effective supervision. Performance gradually declines when distillation is applied to later segments, and can even fall below the initial policy. This suggests that teacher feedback may become less reliable along the rollout (see Fig.~\ref{fig:position_kl_curve}), and late segments may introduce noisy or harmful signals. This explains why standard OPD may sometimes be suboptimal:
\begin{idea}{}
Full rollout OPD aggregates gradients over the entire rollout, including unreliable signals from late rollout positions. In contrast, TOPD focuses on reliable prefixes, thereby reducing cost while improving distillation quality.
\end{idea}
\noindent Overall, this analysis supports our key claim that full rollouts are not always necessary for efficient OPD.

\section{Conclusion}
In this work, we study whether full rollouts are necessary for efficient OPD. 
We show that OPD can be inefficient because full rollouts are costly and teacher feedback on late positions may become unreliable. 
Motivated by this observation, we propose two simple horizon-control strategies: POPD, which gradually expands the rollout horizon, and TOPD,  which performs distillation only on truncated rollouts.
Experiments on LLM reasoning and autoregressive control tasks show that POPD improves training efficiency, while TOPD achieves a favorable trade-off between cost and performance. 
These results suggest that full rollouts are not always necessary for OPD, and that controlling the rollout horizon provides a simple and practical way to improve the efficiency of long-horizon policy learning.

\paragraph{Discussion.} Our findings also raise a broader question for RLVR, namely whether training on truncated rollout segments can effectively unlock performance on longer rollouts. 
If early rollout segments already contain sufficient information about the reasoning patterns, then optimizing truncated rollouts may provide a cost-effective way to induce behaviors that generalize to longer-horizon reasoning. 
However, unlike full rollouts, truncated rollouts are typically incomplete and may not directly lead to a verifiable final answer, making their evaluation a central challenge for RLVR. 
Developing reliable credit assignment or proxy evaluation methods for such truncated rollouts may therefore be crucial for extending horizon-control ideas from OPD to RLVR. 
This perspective suggests that horizon control may be useful not only for OPD, but also for improving the efficiency of RLVR, where long-horizon rollouts are often expensive and noisy.

\section{Limitations}
This work focuses on two simple and broadly applicable horizon-control strategies to validate the effectiveness of controlling the rollout horizon. 
A natural next step is to develop adaptive horizon-control mechanisms.
For example, the rollout horizon might be dynamically adjusted based on the KL divergence between teacher and student, student entropy, teacher confidence, or the variance of log-ratio signals, thereby further improving the training efficiency and supervision quality of OPD.

\bibliography{reference}

\clearpage
\appendix

\section{Related Work}
MiniLLM first formalized OPD for LLMs under a reverse KL objective optimized via policy gradient \citep{gu2024minillm,yue2025does}. Unlike offline distillation \citep{kim2016sequencelevelknowledgedistillation}, which aligns the student with teacher outputs or logits on fixed traces, OPD trains on student-generated prefixes and therefore better matches the distribution encountered by the student during generation. This property helps mitigate the distribution mismatch in offline distillation and enables a student model to effectively absorb capabilities from one or multiple teacher models. As a result, OPD has become an attractive component in recent LLM post-training pipelines \citep{yang2025qwen3technicalreport, lu2025onpolicydistillation, coreteam2026mimov2flashtechnicalreport, glm5team2026glm5vibecodingagentic}. Recent studies have further explored how OPD can be combined with RL \citep{yang2026selfdistilledrlvr}, or extended to self-distillation \citep{penaloza2026privilegedinformationdistillationlanguage,shenfeld2026selfdistillationenablescontinuallearning,2026reinforcementlearningselfdistillation,zhang2026piplaymultiagentselfplayprivileged}, where a single model serves as its own teacher by conditioning on privileged information such as expert demonstration \citep{penaloza2026privilegedinformationdistillationlanguage,shenfeld2026selfdistillationenablescontinuallearning}, execution feedback \citep{2026reinforcementlearningselfdistillation, wang2026openclawrltrainagentsimply}, question construction path \citep{zhang2026piplaymultiagentselfplayprivileged}, or other high-quality auxiliary signals. Despite this progress, the efficiency of OPD remains relatively underexplored. Existing efforts such as EffOPD \citep{cai2026learningforeseeunveilingunlocking} analyze efficient OPD from the perspective of parameter dynamics and optimization behavior. In contrast, our work studies OPD efficiency from the perspective of rollout horizon control. We show that full rollouts are not always necessary for effective OPD, and that prioritizing reliable rollout segments can substantially reduce training cost while preserving strong distillation performance. This direction is orthogonal to parameter dynamics based approaches.

\section{Implementation}
\label{app:implementation}
In the LLM reasoning experiments, we set $\Delta H=9$ and ${\Delta k}=1$. 
For all training runs, we set the original maximum response length to the value used in the teacher's training configuration, specifically 15360 for JustRL-R1-1.5B and JustRL-Nemotron-1.5B.
In the autoregressive navigation task, we set $\Delta H=1$ and ${\Delta k}=2$. 
For the simple navigation task and the autoregressive navigation task, the original maximum rollout horizons are set to $50$.
The remaining hyperparameters for the LLM reasoning experiments are reported in Table~\ref{tab:llm_Hyperparameters}. 
The remaining parameters for the simple navigation task and the autoregressive navigation task are shown in Table~\ref{tab:mlp_policy} and Table~\ref{tab:transformer_policy}, respectively.

\begin{table}[h]
\centering
\caption{\textbf{Hyperparameters for OPD}}
\renewcommand{\arraystretch}{1.15}
\begin{tabular}{lc}
\toprule
Training temperature & \texttt{1.0} \\
Global batch size & \texttt{64} \\
Mini batch size & \texttt{64} \\
Rollout number & \texttt{4} \\
LogProb top-$K$ & \texttt{16} \\
Top-$K$ strategy & \texttt{Student Top-$K$} \\
Top-$p$ & \texttt{1.0} \\
Max prompt length & \texttt{1024} \\
Max response length & \texttt{15360} \\
Learning rate & \texttt{1e-6} \\
Epoch & \texttt{2} \\
KL Coefficient & \texttt{0.0} \\
\bottomrule
\end{tabular}
\label{tab:llm_Hyperparameters}
\end{table}

\begin{table}[h]
\centering
\caption{\textbf{Architecture of the MLP policy network used in the simple navigation task.}}
\renewcommand{\arraystretch}{1.15}
\begin{tabular}{lc}
\toprule
Policy network & \texttt{MLP} \\
Hidden layers & \texttt{3} \\
Hidden dimension & \texttt{64} \\
Activation function & \texttt{Tanh} \\
Mean head & \texttt{Linear(64, 2)} \\
Log-std head & \texttt{Linear(64, 2)} \\
Output distribution & \texttt{2D Gaussian} \\
Action scale & \texttt{0.8} \\
\bottomrule
\end{tabular}
\label{tab:mlp_policy}
\end{table}

\begin{table}[h]
\centering
\caption{\textbf{Architecture of the Transformer policy network used in the autoregressive navigation task.}}
\renewcommand{\arraystretch}{1.15}
\begin{tabular}{lc}
\toprule
Hidden dimension & \texttt{64} \\
Attention heads & \texttt{4} \\
Encoder layers & \texttt{2} \\
Decoder layers & \texttt{2} \\
Activation function & \texttt{GELU} \\
Dropout & \texttt{0.0} \\
Normalization & \texttt{Pre-LN} \\
Output distribution & \texttt{2D Gaussian} \\
Mean head & \texttt{Linear(64, 2)} \\
Log-std head & \texttt{Linear(64, 2)} \\
Acceleration scale & \texttt{0.1} \\
\bottomrule
\end{tabular}
\label{tab:transformer_policy}
\end{table}

\section{Derivation of Future-Noise Accumulation}
\label{app:future_noise_derivation}

\subsection{Noise Accumulation}

Let the ideal sequence-level advantage be
\begin{equation}
\small
    A^{\star}_t
    =
    \sum_{k=t}^{T}
    \gamma^{k-t} r_k^\star,
\end{equation}
where $r_k^\star$ denotes the ideal log-ratio signal, and we assume that the actual received OPD signal can be decomposed as 
\begin{equation}
\small
    r_k
    =
    r_k^\star + \sigma_k z_k,
    \qquad
    z_k \sim \mathcal{N}(0,1),
\end{equation}
where $\sigma_k z_k$ is a stochastic noise term. Then
{\small
\begin{align}
    \hat{A}^{\mathrm{seq}}_t - A^{\star}_t
    &=
    \sum_{k=t}^{T}
    \gamma^{k-t}
    \left(
    r_k - r_k^\star
    \right) \\
    &=
    \sum_{k=t}^{T}
    \gamma^{k-t} \sigma_k z_k.
\end{align}}
Assuming the noise variables $\{z_k\}_{k=t}^{T}$ are independent, then the mean-squared error of the sequence-level advantage satisfies
\begin{equation}
\small
\mathrm{MSE}_t 
    \left(
    \hat{A}^{\mathrm{seq}}_t - A^{\star}_t
    \right)
    =
    \sum_{k=t}^{T} \gamma^{2(k-t)} \sigma_k^2.
\end{equation}
When $\gamma=1$ and the noise scale grows with the rollout position, e.g., $\sigma_k=\delta k$, the mean-squared error becomes
\begin{equation}
\small
    \mathrm{MSE}_t 
    \left(
    \hat{A}^{\mathrm{seq}}_t - A^{\star}_t
    \right)
    =
    \delta^2
    \sum_{k=t}^{T} k^2.
\end{equation}
Using the square-sum formula,
\begin{equation}
\small
    \sum_{k=t}^{T} k^2
    =
    \frac{
    T(T+1)(2T+1)
    -
    (t-1)t(2t-1)
    }{6},
\end{equation}
we obtain
\begin{equation}
\small
\begin{aligned}
    \mathrm{MSE}_t &
    \left(
    \hat{A}^{\mathrm{seq}}_t - A^{\star}_t
    \right)
    = \\&
    \delta^2
    \frac{
    T(T+1)(2T+1)
    -
    (t-1)t(2t-1)
    }{6}.
\end{aligned}
\end{equation}
In particular, for the first token $t=1$,
\begin{equation}
\small
\begin{aligned}
    \mathrm{MSE}_t 
    \left(
    \hat{A}^{\mathrm{seq}}_1 - A^{\star}_1
    \right)
    = 
    \delta^2
    \frac{T(T+1)(2T+1)}{6}
    =
    O(\delta^2 T^3).
\end{aligned}
\end{equation}
This shows that early-token gradients in sequence-level OPD can accumulate substantial future noise in long-horizon rollouts.

\section{Autoregressive Navigation Task}
\label{app:autoregressive_navigation_task}
\begin{promptbox}{Autoregressive Navigation Task}
We consider a more challenging 2D control task compared with the simple navigation task (Sec~\ref{Degraded_Reliability}).
The teacher policy is trained with REINFORCE toward a predefined teacher target, while the student initial policy is independently trained with REINFORCE toward a distinct student initial target. 
Both policies are parameterized by a lightweight Transformer network \citep{NIPS2017_3f5ee243}.

\vspace{0.5em}

At each step, the state is represented as 
$s_t=[\mathrm{task\ id}, \mathrm{velocity\ history}, \mathrm{time\ sequence}]$. 
The velocity history records the previous velocity sequence, and the time sequence provides temporal information for autoregressive decision making.

\vspace{0.5em}

The action $a_t=[a_x,a_y]$ specifies the acceleration along the two spatial dimensions. 
The velocity transition follows
\vspace{-0.3em}
$$
\small
v_{t+1}=v_t+a_t+z,\quad z\sim\mathcal{N}(0,\sigma),
$$
\end{promptbox}

\section{Further Analysis }
\subsection{Sequence-Level OPD Under Optimal Teacher Policy}
\label{app:Optimal_Teacher_Policy_simple_task}
Proposition~\ref{prop:future_noise} explains why sequence-level OPD is sensitive to long horizons. Even if the log-ratio signal at the current token is reliable, the sequence-level advantage may still be corrupted by noisy future signals. This explains the empirical observation in Fig.~\ref{fig:two_by_five}.  Under a small policy mismatch, future signals remain useful and both token-level and sequence-level OPD variants work well. Under a large mismatch, the future signals become unreliable, and sequence-level OPD can be substantially worse than token-level OPD.

\begin{figure}[h]
    \centering

    \includegraphics[width=\linewidth]{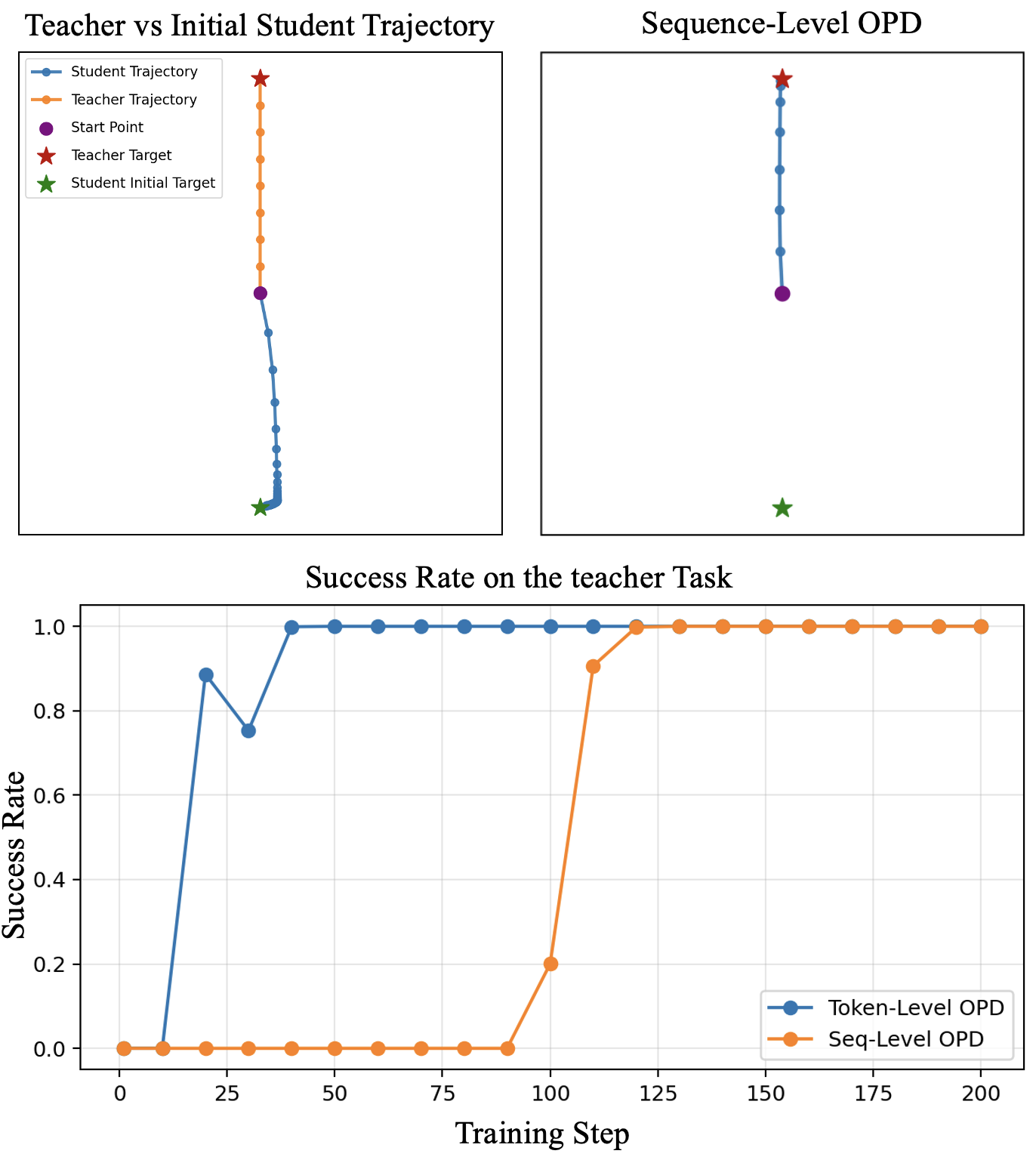}
    \caption{
    \textbf{Sequence-level OPD under an analytic optimal teacher policy.}
    When the teacher is optimal, future log-ratio signals remain reliable despite student mismatch, allowing sequence-level OPD to learn an accurate policy.
    However, sequence-level OPD still aggregates future signals into updates of early tokens, which may introduce additional variance and lead to slower convergence than token-level OPD \citep{fu2026revisitingonpolicydistillationempirical}.
    }
    \label{fig:speed_compare180}
\end{figure}

It is worth noting that this failure mode is largely mitigated when the teacher policy is optimal (Fig.~\ref{fig:speed_compare180}). In this case, even if the initial student policy is substantially different from the teacher policy, the future log-ratio signals are still generated from an accurate policy rather than from an unreliable or biased teacher. Therefore, the accumulated future terms in sequence-level OPD can still provide a correct learning direction and guide the student toward the optimal policy. However, the return-to-go form of sequence-level OPD still aggregates multiple future log-ratio terms into the update of each early token. Therefore, although the systematic bias is removed, the variance introduced by the accumulation of log-ratio signals remains. This explains why sequence-level OPD can still learn an accurate policy under an optimal teacher, but its convergence may be slightly slower than token-level OPD, which relies only on the log-ratio of the current token and therefore avoids the variance induced by log-ratio accumulation.

\begin{table*}[t]
\centering
\small
\begin{tabular}{lcccc}
\toprule
Method & $\rho$ & Response length & Effective length & Memory ratio \\
\midrule
OPD & $1.00$ & $15360$ & $16384$ & $100.00\%$ \\
TOPD & $0.50$ & $7680$  & $8704$  & $53.13\%$  \\
TOPD & $0.25$ & $3840$  & $4864$  & $29.69\%$  \\
TOPD & $0.10$ & $1536$  & $2560$  & $15.63\%$  \\
\bottomrule
\end{tabular}
\caption{
Estimated length-dependent memory ratio of TOPD compared with standard OPD.
The ratio is computed as $(P+\rho T)/(P+T)$ with $P=1024$ and $T=15360$.
Although the response-dependent memory scales approximately with $\rho$, the prompt tokens remain unchanged, so the total length-dependent memory ratio is larger than $\rho$.
}
\label{tab:topd_memory_ratio}
\end{table*}

\section{Memory Analysis of TOPD}
\label{app:memory_analysis}

We provide a simple memory analysis to clarify why TOPD reduces the GPU memory cost of OPD.
Let $P$ denote the prompt length, $T$ denote the maximum response length used by standard OPD, and $H=\rho T$ denote the truncated rollout horizon used by TOPD.
Standard OPD generates and distills the full response of length $T$, while TOPD only generates and distills the first $H$ response tokens.

The peak memory consumption of OPD can be decomposed into two parts:
\begin{equation}
M_{\mathrm{peak}}
=
M_{\mathrm{static}}
+
M_{\mathrm{len}},
\end{equation}
where $M_{\mathrm{static}}$ denotes length-independent memory, including model parameters, gradients, and optimizer states.
This term is unchanged between OPD and TOPD.
In contrast, $M_{\mathrm{len}}$ denotes length-dependent memory, including generation KV cache, log-probability buffers, and training activations. These terms scale with the effective sequence length.

For standard OPD, the sequence length is $L_{\mathrm{OPD}} = P + T$.
For TOPD, the effective sequence length becomes $L_{\mathrm{TOPD}} = P + H = P + \rho T$.
Therefore, the ratio of length-dependent memory between TOPD and full OPD is approximately
\begin{equation}
\frac{M_{\mathrm{len}}^{\mathrm{TOPD}}}
{M_{\mathrm{len}}^{\mathrm{OPD}}}
\approx
\frac{P+\rho T}{P+T}.
\label{eq:topd_memory_ratio}
\end{equation}
If we only consider the response part, the ratio further simplifies to
\begin{equation}
\frac{H}{T}=\rho.
\end{equation}
Thus, TOPD reduces the memory of the response part approximately in proportion to the truncation ratio $\rho$. For example, when we use a maximum prompt length of $P=1024$ and a maximum response length of $T=15360$.
The resulting effective sequence lengths and length-dependent memory ratios are summarized in Table~\ref{tab:topd_memory_ratio}.

Finally, the actual peak memory reduction is smaller than the ideal length-dependent reduction because the static memory term is unchanged:
\begin{equation}
\frac{
M_{\mathrm{peak}}^{\mathrm{TOPD}}
}{
M_{\mathrm{peak}}^{\mathrm{OPD}}
}
=
\frac{
M_{\mathrm{static}} + c(P+\rho T)
}{
M_{\mathrm{static}} + c(P+T)
},
\label{eq:peak_memory_ratio}
\end{equation}
where $c$ is a constant that summarizes the memory cost of per token.
Eq.~\eqref{eq:peak_memory_ratio} shows that TOPD mainly reduces the length-dependent component of OPD memory, while memory $M_{\mathrm{static}}$ remains the same.
Therefore, TOPD does not necessarily reduce the measured peak GPU memory by exactly $\rho$, but it substantially lowers the memory associated with long rollouts, especially KV cache and log-probability computation buffers.

\end{document}